# Large Language Model Agent as a Mechanical Designer

**Yayati Jadhav**
Mechanical Engineering,
Carnegie Mellon University,
5000 Forbes Ave,
Pittsburgh, PA, USA
email: yayatij@andrew.cmu.edu

**Amir Barati Farimani** [1]
Mechanical Engineering,
Carnegie Mellon University,
5000 Forbes Ave,
Pittsburgh, PA, USA
email: barati@cmu.edu

*Conventional mechanical design paradigms follow an iterative refinement process where initial concepts evolve through cycles of expertise-driven, heuristic-informed assessment guided by resource-intensive Finite Element Method (FEM) analyses to achieve performance specifications. While numerous machine learning models have been developed to tackle various aspects of the expert-driven, iterative design process, however, most require large training datasets and significant computational resources. Furthermore, deep learning-based approaches are often tailored to specific domains and tasks, which limits their generalizability to broader design problems. This creates a trade-off between the scalability of such methods and the flexibility needed for diverse design applications.*

*To address this, we propose a framework that leverages a pretrained Large Language Model (LLM) coupled with FEM module to autonomously generate, evaluate, and refine structural designs based on performance specifications and quantitative feedback. The LLM operates without domain-specific fine-tuning, relying solely on general reasoning to iteratively propose candidate structures, interpret FEM-derived results, and implement informed modifications grounded in structural mechanics. Demonstrated using 2D truss structures, this approach highlights the LLM's ability to navigate complex design spaces, balance competing objectives, and identify appropriate termination points when further optimization yields diminishing returns. Comparative analysis against traditional optimization methods, such as Non-dominated Sorting Genetic Algorithm II (NSGA-II), shows that LLM-guided optimization achieves faster convergence and requires fewer FEM evaluations in highly discrete, multi-faceted design spaces characterized by dynamic node generation and discrete member sizing. Experiments across multiple temperature settings (0.5, 1.0, 1.2) and model scales (GPT-4.1 and GPT-4.1-mini) reveal that smaller, distilled models achieve marginally superior constraint satisfaction with fewer iterations, while lower temperatures yield more consistent structural performance. These results highlight the promise of LLM agents to function as a new class of natural language-based, reasoning-driven optimizers capable of independently generating and iteratively improving viable engineering solutions through structured feedback.*

## 1 Introduction

Mechanical design is fundamentally an iterative process, where designers refine initial concepts through a sequence of informed decisions aimed at meeting performance targets. Traditionally, this process relies on expert intuition, heuristic reasoning, and frequent evaluation through resource-intensive Finite Element Method (FEM) simulations [1–5]. Designers must often navigate incomplete problem definitions, evolving requirements, conflicting objectives, and strict time constraints, where nonlinear material behavior and complex coupling between design variables make improvement in one metric (e.g., mass) potentially detrimental to another (e.g., stress) [6–9].

This complexity spans a wide spectrum of mechanical design activities, from early-stage concept generation, mechanism synthesis, and configuration design to detailed optimization of structural components and material layouts [10, 11]. Tasks such as identifying functional architectures, arranging components, optimizing kinematic behavior, and ensuring manufacturability require designers to reason across multiple abstraction levels and balance qualitative and quantitative trade-offs. Addressing such broad, heterogeneous, and constraint-rich design spaces remains a major challenge for computational design automation frameworks.

Various computational strategies have been developed to address specific aspects of the mechanical design process, particularly in structural and material layout optimization. Parametric frameworks such as Solid Isotropic Material with Penalization (SIMP) [12, 13], often coupled with optimization algorithms like the Optimality Criteria (OC) method or the Method of Moving Asymptotes (MMA), have been widely adopted for topology optimization tasks. Gradient-free methods, including Genetic Algorithms (GAs) [14–16], provide greater flexibility in exploring discrete and nonconvex design spaces, albeit at the cost of higher computational demands. More recent approaches like Bidirectional Evolutionary Structural Optimization (BESO) [17, 18] and Variational Topology Optimization (VARTOP) [19, 20] offer fine-grained control over material removal and addition processes to incrementally evolve optimal structures.

While these methods have significantly advanced computational design, each carries inherent limitations. Gradient-based frameworks like SIMP are prone to converging to local minima and exhibit sensitivity to initial conditions. Genetic Algorithms, though more global in search, often require large numbers of evaluations to achieve convergence, making them computationally expensive. Methods such as BESO and VARTOP provide greater control over topology evolution but frequently demand careful parameter tuning and can struggle to resolve fine-scale geometric features. Critically, all of these approaches rely heavily on repeated FEM evaluations for performance assessment [21], leading to substantial computational overhead that limits their scalability in high-fidelity or early-stage design scenarios where rapid iteration is essential.

Data-driven deep learning (DL) methods have enabled efficient data processing and distillation of relevant information from high-dimensional datasets [22–24], leading to highly accurate models in fields like additive manufacturing [25, 26], solid mechanics [27–30], among others [31, 32]. In recent years these methods have shifted mechanical design from an iterative physics-based approach to a more dynamic, data-driven approach.

[1] Corresponding Author.
Version 1.18, April 30, 2025

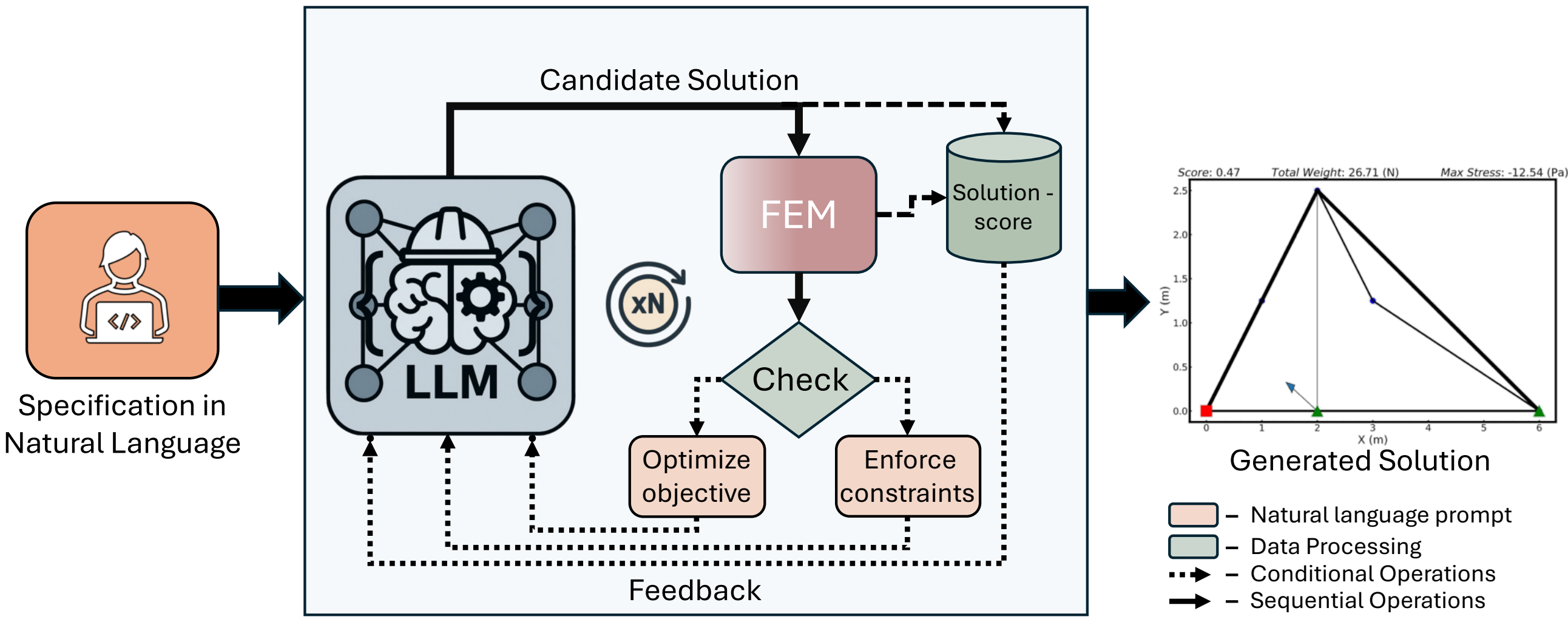

**Fig. 1 Overview of the proposed closed-loop optimization framework integrating a pretrained Large Language Model (LLM) with a Finite Element Method (FEM) module. A user prompt describing structural goals, loads, and boundary conditions is processed by the LLM to generate a candidate truss design. The FEM module evaluates the structure's performance, and a *Check* module determines whether the constraints are satisfied. If feasible, the LLM is prompted to optimize further; if not, it is redirected with a constraint-focused prompt to correct violations. Each solution–score pair is appended to a ranked history of the previous $N-1$ iterations, which is reinjected into the LLM as structured feedback. This archive enables the LLM to reason over past designs, identify trends, and iteratively improve solutions based on both constraint satisfaction and objective performance.**

In recent years, the intersection of deep learning and mechanical design has seen remarkable progress, transforming the conceptualization and realization of 3D structures. This advancement has facilitated the creation of 3D models across a diverse array of formats, including voxels [33, 34], point clouds [35–37], neural radiance fields (NeRF) [38, 39], boundary representation (B-rep) [40, 41], triangular meshes [42, 43], and computer-aided design (CAD) operation sequence [44]. The capability to generate 3D structures from diverse input modalities, including text [45, 46], images [47], and sketches [48–50], has expanded the possibilities and made the field of mechanical design more accessible and versatile. However, a notable limitation is that these models are not inherently designed to account for mechanical specifications or functional constraints.

Integrating deep learning into mechanical design workflows marks a substantial shift in how structural optimization is approached, especially given the limitations of conventional CAD-based modeling and physics-based optimization methods. One emerging direction frames optimal structure generation as a data-driven learning task, similar to image segmentation, where material distribution is optimized to satisfy performance constraints [51]. Another approach uses generative models such as conditional generative adversarial networks (cGANs) [52, 53] and diffusion models [54], trained on large datasets of high-performing designs to produce new, potentially optimal structures. Despite their creative potential, these models are inherently stochastic and require validation through finite element analysis (FEA) to ensure feasibility. Deep learning also enables the development of surrogate models that predict mechanical responses such as stress and strain fields, significantly reducing the computational cost of repeated simulations [30, 55, 56]. These surrogates can be integrated into broader optimization workflows [57–59]. Another promising strategy combines neural networks with metaheuristic search algorithms like genetic algorithms, where the neural network approximates the structure's physical behavior and the search algorithm explores the design space for optimal configurations [60].

While deep learning offers remarkable capabilities, it faces notable challenges and limitations. Notably, generative models like cGANs [52, 55] and diffusion models [30, 54] often require vast amounts of diverse datasets to effectively model the dynamics of underlying systems [61–63]. This is particularly challenging in contexts like FEM simulations, where producing high-fidelity, fine-mesh datasets demands considerable computational resources. Additionally, deep learning models are generally task-specific; they excel in the tasks they are trained for but struggle to adapt to significantly different tasks or scenarios, highlighting a lack of flexibility compared to human problem-solving abilities.

Despite these advances, there remains a critical need for design frameworks that combine flexible reasoning, rapid adaptation to evolving objectives, and efficient exploration of complex, constraint-rich spaces, without incurring the high overhead of task-specific model tuning or dataset generation. In practical mechanical design settings, input and output modalities are highly diverse: initial concepts may be expressed as sketches, textual specifications, CAD models, boundary conditions, or performance targets, while outputs may range from topology layouts and component arrangements to full manufacturable models. Evaluation formats are equally heterogeneous, spanning analytical approximations, high-fidelity numerical simulations, and empirical testing of prototypes. Compounding this complexity, many design tasks involve mixed-variable optimization problems, where continuous parameters (e.g., nodal coordinates, thicknesses) interact with discrete or categorical choices (e.g., member connectivity, material selections, component topologies). Traditional optimization frameworks, whether gradient-based or gradient-free, often struggle to simultaneously accommodate this diversity of inputs, outputs, and evaluation mechanisms, particularly under dynamic, multi-objective conditions where design criteria evolve. As mechanical systems grow in complexity and the demand for early-stage exploration increases, there is a pressing need for reasoning-driven approaches capable of flexibly integrating diverse information streams, balancing multiple objectives, and making coherent, iterative design modifications in response to structured but non-differentiable feedback.

Large Language Models (LLMs), powered by transformer ar-

chitecture [64, 65] and trained on extensive datasets [66–71], have achieved remarkable progress in various natural language processing (NLP) tasks. These tasks include text generation [72], compliance with specific task directives [73, 74], and the demonstration of emergent reasoning capabilities [75, 76]. Given the significant computational resources required for training and fine-tuning LLMs for specific downstream tasks, these models have demonstrated an exceptional ability to generalize to new tasks and domains. This adaptability is achieved through the in-context learning (ICL) paradigm [66], which has significantly deepened our understanding of LLMs' capabilities. By utilizing minimal natural language templates and requiring no extensive fine-tuning, LLMs have established themselves as efficient "few-shot learners" [77, 78].

Large Language Models (LLMs) have rapidly proven their versatility across scientific domains by adapting to new contexts and learning on the fly. In chemistry, LLMs autonomously design, plan, and execute intricate experiments [79, 80]. In mathematics and computer science, they've devised novel solutions to problems like the cap set problem and created more efficient algorithms for challenges such as bin-packing [81]. Their influence extends through biomedical research [82–84], materials science [85, 86], and environmental studies [87], among other fields [88–91]. Beyond these applications, LLMs have even matched or outperformed bespoke heuristics on foundational optimization tasks—like linear regression and the traveling-salesman problem—using only simple prompts [92].

In mechanical engineering specifically, fine-tuned LLMs have excelled at knowledge retrieval, hypothesis generation, and agent-based modeling via ontological knowledge graphs [93]. They've automated early-stage concept synthesis [94, 95] and tackled concrete design challenges—including sketch similarity analysis, material selection, engineering drawing interpretation, CAD generation, and spatial reasoning—often rivaling domain-specific tools [96].

Our work builds on these successes by extending LLM-driven optimization from "white-box" objectives like algebraic functions to truly "black-box" evaluations requiring spatial reasoning and implicit engineering judgment. We embed a pre-trained LLM in a closed-loop pipeline with a FEM module that scores each design iteration. Without any further fine-tuning, the LLM uses the FEM feedback to propose, assess, and refine 2D truss geometries under loads and supports, demonstrating the model's ability to internalize domain knowledge and navigate complex, multidimensional design spaces.

In this study, we introduce a framework that combines in-context learning with few-shot prompt templates, harnessing the inherent reasoning and optimization capabilities of LLMs for structural design. Through our truss design case study, we show how LLM agents can generate initial concepts, iteratively improve them, and handle both mixed-numeric (mix of discrete and continuous space) and categorical data, surpassing traditional optimizers that typically only process continuous numeric inputs. This work underscores the promise of LLM-based agents to autonomously develop and implement robust engineering solutions through structured, feedback-driven learning.

As shown in Fig. 1, the proposed framework begins with structural requirements expressed in natural language, which a pre-trained LLM interprets to generate an initial truss design. The design is evaluated using the FEM module, and feedback—including constraint satisfaction status, optimization goals, and a ranked history of previous solution score pairs—is returned to the LLM as a meta prompt. This enables iterative refinement or new design generation. The ranked history guides the LLM's reasoning by highlighting successful patterns and supporting a balance between exploiting known good designs and exploring novel ones. The process terminates when the LLM detects diminishing performance gains or consistent constraint satisfaction, using its internal reasoning rather than fixed thresholds to conclude optimization.

## 2 Methodology

**2.1 Problem Description.** In this study, we evaluate the capability LLMs to autonomously generate optimized truss structures, capturing a fundamental workflow in mechanical engineering: the creation of an initial structural design, evaluation of its performance, analysis of its behavior under prescribed loading conditions, and iterative refinement to satisfy stringent design and functional requirements.

To systematically assess the capabilities of LLMs, a framework comprising two primary tasks, each with three specification variations, is established. The evaluation spans a range of constraint difficulties, enabling a systematic investigation of the model's ability to follow instructions, satisfy complex design requirements, optimize structural configurations, and reason across successive design iterations. This framework facilitates the analysis of the LLM's capacity for solution exploration, iterative refinement, and strategic modifications informed by prior outcomes.

The problem formulation is common to both tasks. Given a set of predefined support nodes $S$, load nodes $L$, and a cross-sectional area dictionary $\mathcal{A}$ specifying the available member areas, the LLM must synthesize an optimized truss structure by autonomously:

- Generating additional node placements $N \subset \mathbb{R}^2$,
- Defining the connectivity between nodes as a set of edges $E \subset (S \cup L \cup N) \times (S \cup L \cup N)$,
- Assigning cross-sectional areas $A_e \in \mathcal{A}$ to each member $e \in E$.

The model must achieve task-specific optimization objectives while adhering to all specified constraints.

**Task 1:** The LLM must generate a structure that satisfies maximum allowable stress constraints $|\sigma_e| \leq \sigma_{\max}$ and a maximum allowable mass constraint $M_{\text{total}} \leq M_{\max}$. Once these baseline constraints are satisfied, the model further minimizes the maximum observed stress according to the following objective:

$$\min_{E, A_e, N} \max_{e \in E} |\sigma_e|$$

The LLM autonomously determines when to terminate the optimization, balancing incremental improvements against diminishing returns from continued refinement.

**Task 2:** The LLM must design a structure that minimizes the stress-to-weight ratio, defined as:

$$\text{Stress-to-Weight Ratio (swr)} = \frac{\max(|\sigma_e|)}{\sum_{e \in E}(L_e \times A_e)}$$

subject to the constraint $M_{\text{total}} \leq M_{\max}$. This task requires the model to navigate complex trade-offs between reducing structural mass and maintaining mechanical robustness.

In both tasks, the LLM is granted the flexibility to add nodes at arbitrary locations, define strategic connections, and select cross-sectional areas from the available set. Successful design generation depends on effectively balancing exploration of novel structural possibilities with exploitation of established configurations to satisfy performance and constraint requirements. Navigating this trade-off is essential for producing structurally efficient and compliant structures.

Table 1 summarizes the experimental setup for each task and its variations, specifying the initial conditions, including nodal locations for load and supports. Task 1 explores three variations with progressively stricter maximum stress limits, while Task 2 targets different stress-to-weight ratios, ranging from low to high. This setup highlights the range of specifications used to evaluate the LLM agent's adaptability to varying constraints and objectives.

To further investigate the stochastic behavior of LLMs, we varied the temperature parameter across three settings: 0.5, 1.0, and 1.2. The temperature parameter controls the randomness of token sampling during generation; lower values result in more deterministic and focused outputs, whereas higher values encourage greater diversity at the expense of consistency. Each experiment was repeated ten times for each temperature setting to systematically evaluate how sampling variability influences the model's ability to generate constraint-compliant and optimized truss designs.

| | Node Position (Node, (x,y)) | Load Position (Node, (magnitude, direction)) | Support Positions (Node, support type) | Variation No. | Specifications 1 | Specifications 2 (Total weight of structure in N) | Cross-section ID of members to be used |
|---|---|---|---|---|---|---|---|
| Task 1 | "node_1": (0, 0), "node_2": (6, 0), "node_3": (2, 0), | "node_3": (-10, -45) | "node_1": "pinned", "node_2": "roller", | 1 | Maximum stress: 15 (Pa) | 30 | "0": 1, "1": 0.195, "2": 0.782, "3": 1.759, "4": 3.128, "5": 4.887, "6": 7.037, "7": 9.578, "8": 12.511, "9": 15.834, "10": 19.548 |
| | | | | 2 | Maximum stress : 20 (Pa) | 30 | |
| | | | | 3 | Maximum stress: 30 (Pa) | 30 | |
| Task 2 | "node_1": (0, 0), "node_2": (6, 0), "node_3": (2, 0), | "node_3": (-15, -30) | "node_1": "pinned", "node_2": "roller", "node_3": "roller", | 1 | Stress/Weight ratio : 0.5 | 30 | |
| | | | | 2 | Stress/Weight : 0.75 | 30 | |
| | | | | 3 | Stress/Weight : 1 | 30 | |

**Table 1 Truss Design Specifications. Variations differ by target stress or stress-to-weight ratios, while available cross-sectional areas remain grouped by task.**

In addition, to study the effect of model scale, experiments were conducted using different versions of GPT-4.1: the standard GPT-4.1 model and the smaller GPT-4.1-mini. Attempts were also made to evaluate performance on GPT-4.1-nano; however, the quality of outputs at this scale was consistently poor and failed to meet the minimum requirements for meaningful evaluation. This extended analysis provides a comprehensive view of how both stochasticity and model capacity influence the robustness and reliability of LLM-driven structural design generation.

**2.2 Prompt Design.** Prompt design plays a pivotal role in effectively harnessing the capabilities of LLMs, acting as the critical interface between user intent and model interpretation. It involves the strategic construction of input queries, or "prompts," to direct the model toward producing outputs that are accurate, relevant, and contextually appropriate [97]. In addition to guiding generation, effective prompt design is essential for extracting targeted information from the model, ensuring that the responses align closely with the task objectives. The performance of LLM agents in specialized tasks is highly sensitive to prompt precision and structure; minor variations can substantially impact output quality, consistency, and relevance. Consequently, prompt engineering has become a fundamental technique for optimizing LLM-driven workflows, enabling users to steer model behavior reliably and achieve desired outcomes across diverse applications [98, 99].

The design of effective prompts is particularly critical in the context of structural generation and optimization, especially when structured outputs are required for downstream tasks. To ensure consistency and usability across multiple stages of the workflow, prompts must generate outputs that are both machine-interpretable and aligned with engineering objectives.

The initial prompt (Fig.2) to generate the first structure was organized into several key sections, each addressing a specific aspect of the task. The problem was framed as a single-shot optimization, requiring the LLM to produce a complete solution without reliance on prior designs, thereby encouraging comprehensive internal reasoning before output generation.

Inputs provided to the model included node positions (`node_dict`), applied loads (`load`), support information (`supports`), and available cross-sectional areas (`area_id`), collectively defining the design space and boundary conditions. The objective was to minimize the maximum stress in the truss members while satisfying predefined constraints, specifically a stress limit of ±`max_stress` and a total mass limit of `max_weight`.

Detailed instructions specified that the model must generate a fully connected 2D truss structure without modifying load or support nodes. Additional nodes could be introduced with precise

***Status:***
Treat this as a single-shot optimization task. Minimize stress while ensuring the generated closed (all nodes connected) 2D Truss remains within the constraints.

***Inputs:***
- Node_dict: Given node position: *{node_dict}*
- Loads: Forces acting on node (node_id: (magnitude, direction)): *{load}*
- Supports: Support information *{supports}*
- area_id: Area ID values to choose cross section from : *{area_id}*

***Objectives:***
1. Generate optimal closed truss structure that minimizes stress, while satisfying all given constraints.

***Constraints:***
1. Max stress (compressive(-) and tensile(+)) <= ±*{max_stress}*
2. Total mass: Mass (sum of member length × area from *{area_id}*) <= *{max_weight}*

***Instructions:***
1. Treat this as a single-shot optimization task. No prior designs exist.
2. Design a fully connected (closed) 2D truss — all nodes must be part of a single closed connected structure.
3. Your goal is to minimize the stress while strictly satisfying all constraints.
4. Use only the given nodes, loads, supports, and area IDs. Do not modify load and support node positions.
5. Generate new nodes (if needed) with precise float values, Do not modify load and support node positions.
6. Choose area_id values only from *{area_id}* using string keys.
7. Avoid duplicate nodes or member connections. All members must reference existing, unique nodes.
8. Use your understanding of structural mechanics — infer effective strategies (e.g., triangulation, symmetry, direct load paths) to create most optimal structure.
9. Estimate stress and mass using basic structural principles (e.g., static equilibrium, trigonometry). Clearly state any assumptions made.
10. Output must be a complete, valid Python dictionary. All entries must be internally consistent and ready for parsing.

***Output format (python dictionary):***
Return a single valid Python dictionary in the format below. Keep all string fields concise. Use floats for coordinates.
```
{{
  "preamble" :  "Briefly describe the problem, current design status, and the optimization objective.",

  "scratch_pad_1": "Initial interpretation of given structure objectives and constraints. This is the first iteration",

  "analysis": "Detailed analysis of the structural problem based on inputs. Include interpretation of load paths, boundary conditions, possible failure modes, and efficiency of current design",

  "scratch_pad_2": "Reflect on the analysis. Note any patterns, areas of concern, or insights that may guide the design.",

  "reasoning": "Use design principles (e.g., symmetry, triangulation, load distribution) to justify the design direction. Propose structural changes or node repositions to improve performance.",

  "scratch_pad_3": "Outline key assumptions, set-up equations (e.g., force balance, mass calculation, trigonometric equations etc.), and calculations to be performed. Include considerations for member selection or node movement.",

  "calculation": ""Perform required calculations such as updated member forces, stress, and structure mass. Quantify constraint violations if any, and iterate on modifications if needed",
  "scratch_pad_final": "Final scratch pad before you generate the structure verify adherence to objective, constraints and Instructions. Perform modifications to ensure the design meets all requirements.",

  "node_dict": {{
    # Node coordinates must be precise floats based on structural calculations.
    # Each node entry should include design reason,  that will help future optimization.

    "node_1": (x, y), # thought, reason, design choice, calculation
    "node_2": (x, y), # thought, reason, design choice, calculation
  }},
  "member_dict": {example_member},
}}
```

**Fig. 2 Initial prompt used for generating the truss solution. The prompt defines the task status, inputs, objectives, constraints, instructions, and output format to guide the design process.**

floating-point coordinates to optimize performance, but duplicate nodes and redundant connections were prohibited. Cross-sectional areas had to be selected strictly from the provided `area_id` set, with design strategies such as triangulation, symmetry, and direct load paths encouraged to enhance structural efficiency.

The output format was required to follow a strict Python dictionary structure containing `node_dict` and `member_dict`, ensuring direct compatibility with FEM workflows. Each node entry included its coordinates alongside a concise rationale for its placement, supporting both traceability and subsequent evaluation.

To promote logical consistency and deepen reasoning, the prompt incorporated a sequenced scratchpad system. The model was guided through distinct stages: interpreting the task, analyzing boundary conditions, formulating design strategies, setting up calculations, and finally verifying constraint satisfaction before generating the final structure. This structured decomposition leverages the causal attention mechanism inherent to generative pretrained transformers, ensuring that critical task information remains accessible throughout generation. By organizing key decision points sequentially, the prompt maintained model focus on core objectives and constraints, contributing to the generation of coherent, verifiable truss designs.

Once the initial truss structure is generated based on the initial prompt, it is evaluated using FEM to determine whether the design satisfies the prescribed constraints. The evaluation produces a result dictionary containing key performance metrics, including maximum member stress and total structural mass.

If the generated structure satisfies all constraints, the optimization phase proceeds using the optimization prompt (see App.B). This prompt guides the LLM to refine the structure further, aiming to minimize stress while maintaining compliance with all constraints. The model is allowed to fine-tune node placements (excluding nodes with loads or supports) and adjust member connections, utilizing information from the current design, FEM results, and the structure generation history. If no further improvements are feasible within the constraints, the model can decide to terminate optimization by returning the best structure identified.

Conversely, if the initial structure fails to satisfy one or more constraints, the constraint prompt (see App .C) is used. In this case, the model is directed to prioritize achieving feasibility over optimization. The prompt instructs the LLM to analyze the constraint violations, interpret the current design and historical data, and either improve the existing structure, redesign it, or reuse a previously generated solution. Only after achieving a constraint-compliant design does further optimization become relevant.

Both prompts employ a structured scratchpad sequence to support step-by-step reasoning. The model is required to interpret the problem, analyze performance, propose modifications, reason about design choices, perform necessary calculations, and verify constraint satisfaction before finalizing the output. This structured decomposition improves logical consistency and aligns with the causal attention mechanism of generative pretrained transformers, ensuring that critical task information remains available throughout the generation process.

**2.3 Prompt Engineering.** Prompt engineering, also referred to as in-context prompting, involves communicating with LLMs to guide their behavior toward desired outcomes without modifying model weights (fine-tuning) [100]. This approach is distinct from prompt design, which focuses on crafting initial inputs that leverage a model's existing knowledge to optimize performance on specific tasks. Various methods have been proposed to optimize LLM outputs through the strategic selection of in-context examples, including semantic clustering with K-NN [101], constructing directed graphs based on embedding similarity [102], and reinforcement learning techniques such as Q-learning for sample selection [103], among others [104–106].

However, these example selection strategies can introduce biases by favoring samples similar to the training data, which may restrict the model's ability to explore diverse solutions. This limitation is particularly problematic in design optimization, where broad exploration is critical for fostering innovation and discovering novel solutions.

To enhance exploration, several methods have been developed. Self-consistency sampling generates multiple outputs at nonzero temperatures and selects the best candidate [107], while chain-of-thought prompting encourages step-by-step reasoning to improve answer quality [108, 109]. Tree-of-thought prompting extends this by branching reasoning at each step to explore multiple possible outcomes [110]. Although these techniques promote exploration, they can also significantly increase computational cost and output complexity. For instance, self-consistency requires evaluating multiple outputs, and tree-of-thought reasoning can cause an exponential growth in the solution space.

The ReAct framework mitigates some of these challenges by interleaving reasoning traces with task-specific actions, enabling dynamic interaction with external environments and improving the efficiency of solution generation [111].

Our proposed framework leverages the ReAct mechanism [111] to iteratively generate, evaluate, and refine truss structures using a LLM (GPT-4). Beginning from a base prompt, the LLM proposes a structure by specifying node placements $N$, edge connections $E$, and cross-sectional area assignments $A_e$.

**Algorithm 1** LLM-Guided Truss Design and Optimization (Task 1)

**Input:** Supports $S$, Loads $L$, Area set $\mathcal{A}$, Stress limit $\sigma_{\max}$, Mass limit $M_{\max}$
**Output:** Optimized truss structure $(N, E, A_e)$

**// Legend:**
// $S$ : Support nodes  $L$ : Load nodes  $\mathcal{A}$ : Area options
// $N$ : Generated nodes  $E$ : Edges  $A_e$ : Area of edge $e$
// $\sigma_e$ : Stress in member $e$  $\sigma_{\max}$ : Stress limit
// $M_{\text{total}}$ : Structure mass  $M_{\max}$ : Mass limit
// $\mathcal{H}$ : History  $\mathcal{F}$ : Feasible set  $\mathcal{I}$ : Infeasible set
// $\mathcal{R}$ : Ranked designs  `msg_optim` : Optimization prompt
`msg_constraint` : Constraint correction prompt

**for** each run $j$ **do**
  Initialize base prompt `msg`, history $\mathcal{H} \leftarrow \emptyset$
  **for** each iteration $i$ **do**
    Generate $(N, E, A_e) \leftarrow$ `LLM(msg)`
    FEM evaluation: stresses $\sigma_e$, mass $M_{\text{total}}$
    Save $(N, E, A_e, \sigma_e, M_{\text{total}})$ to $\mathcal{H}$
    Identify feasible set:

$$\mathcal{F} = \{(N, E, A_e) \in \mathcal{H} \mid |\sigma_e| \le \sigma_{\max}, M_{\text{total}} \le M_{\max}\}$$

    Sort $\mathcal{F}$ ascending by $\max(|\sigma_e|)$
    Identify infeasible set $\mathcal{I} = \mathcal{H} \setminus \mathcal{F}$, sort by $\max(|\sigma_e|)$
    Merge $\mathcal{R} = \mathcal{F} \cup \mathcal{I}$
    **if** $(N, E, A_e) \in \mathcal{F}$ **then**
      Update `msg_optim` with history $\mathcal{R}$
    **else**
      Update `msg_constraint` with history $\mathcal{R}$
    **end if**
    **if** Early stopping triggered **then**
      **break**
    **end if**
  **end for**
**end for**
**return** Best feasible $(N, E, A_e)$ in $\mathcal{H}$

At each step, the LLM generates a candidate truss structure with up to 10 attempts allowed to account for failures where the output does not conform to the required format. Each valid structure undergoes FEM analysis to compute key structural metrics, including maximum member stresses $\sigma_e$ and total mass $M_{\text{total}}$. These FEM evaluations serve not only as binary constraint checks but also provide detailed quantitative feedback, measuring how closely the generated structure meets stress and mass requirements.

All generated structures and their associated FEM results are stored in a growing history set $\mathcal{H}$, which captures feasible designs, partially feasible solutions, and failed attempts. A ranked history $\mathcal{R}$ is maintained, prioritizing feasible designs based on minimum maximum stress and ordering infeasible designs by proximity to constraint satisfaction. This structured memory allows the LLM to reason over previous iterations, leveraging patterns of success, incremental improvements, and common failure modes to guide future generations.

At each iteration, the LLM is provided with the current design, the complete design history, and the iteration count (up to a maximum of 25 iterations). Based on the outcome of the FEM evaluation, an adaptive prompting strategy is employed: if the generated structure satisfies all constraints, an optimization prompt (`msg_optim`) directs the model to further minimize stress or improve structural efficiency; if the structure is infeasible, a constraint-correction prompt (`msg_constraint`) instructs the LLM to analyze constraint violations and propose modifications. Throughout the process, the LLM leverages the FEM results and ranked history of prior designs to identify favorable patterns, correct past mistakes, and guide new generations. The model is explicitly tasked with autonomously terminating the optimization process once no further meaningful improvements are achievable.

For Task 2, the optimization strategy is staged: initially focusing on satisfying the constraint $M_{\text{total}} \leq M_{\max}$ and $swr \leq swr_{\max}$, followed by minimizing the stress-to-weight ratio. This sequential focus ensures that both critical performance objectives are systematically addressed. The process continues over a fixed number of iterations or until an early stopping condition is triggered, at which point the best feasible structure found across all iterations is returned.

## 3 Results

This section presents the performance evaluation of the LLM-driven truss structure generation and optimization framework. The results are organized according to the two primary tasks: initial constraint satisfaction and subsequent stress minimization. For each task, performance metrics are reported across different experimental conditions, including varying temperature settings and model scales (GPT-4.1 and GPT-4.1-mini).

The evaluation focuses on few key aspects: (1) the ability of the LLM to generate feasible, constraint-compliant truss structures; (2) the effectiveness of the optimization phase in reducing maximum member stress; and (3) the impact of stochasticity (temperature) and model size on performance consistency and solution quality and (4) the convergence behavior across optimization iterations. Performance statistics are aggregated over ten independent trials for each configuration to ensure robust assessment.

**3.1 Generated Results.** Figure 4 illustrates a representative raw solution generated by the LLM agent at the last step, formatted to simplify parsing and integration into the FEM module for evaluation. The structured output format, which includes preamble, scratchpads, analysis, and calculations, explicitly leverages the causal attention mechanism of the underlying GPT model. By enforcing a sequential, step-by-step reasoning format, the prompt design ensures that the model accumulates and integrates relevant context before determining the final nodal and member parameters.

The raw output shows that the LLM first correctly interprets the optimization objective: minimizing maximum member stress while keeping total mass within limits. Through structured analysis and scratchpads, the model identifies that node 4 offers an opportunity for fine-tuning. It systematically applies engineering design principles such as force redistribution, geometric optimization, and mass reduction to guide its modification strategy. The model justifies its decision by setting up force balance assumptions, estimating effects on stress and mass, and validating that all truss members remain properly connected to maintain structural integrity. This logical flow demonstrates the model's ability to perform targeted optimization while adhering to fundamental principles of mechanical design. Furthermore, the decision-making process is informed by the scratchpad structure, which prevents premature conclusions by requiring the model to reason through possible impacts before finalizing the output. Notably, the model autonomously decides to stop after a minor adjustment when further changes are unlikely to yield significant improvements, demonstrating an implicit understanding of diminishing returns in practical design optimization.

The raw model output highlights the ability fo LLM to autonomously reason about geometric and force redistribution strategies, producing not only a feasible structure but also proactively suggesting minor improvements to further reduce stress while preserving constraint satisfaction. The decision to stop after a small adjustment reflects an awareness of diminishing returns in design refinement, aligning with principles of practical structural optimization.

Fig.3 presents the final optimized truss structures generated by the LLM agent for Tasks 1 and 2. The diversity of the resulting designs demonstrates the model's ability to apply core engineering principles—such as force path optimization, triangulation, and mass minimization—while adapting to varying constraint specifications. In Task 1 (Figure 3a), the model generates lightweight, efficiently triangulated structures that distribute loads symmetrically while minimizing maximum stress. In Task 2 (Figure 3b), where the objective shifts toward optimizing the stress-to-weight ratio, the LLM explores alternative topologies that slightly trade off weight for improved load distribution.

Across both tasks, the designs exhibit effective use of direct load paths, minimal unnecessary members, and clear hierarchical structure typical of optimized truss systems. The range of solutions further highlights the LLM's capability not just to satisfy mechanical constraints, but to reason creatively within the design space to produce multiple structurally valid configurations tailored to different optimization targets.

**3.2 Framework performance.** Across all experiments, the LLM agent consistently generated truss structures that satisfied the specified constraint conditions, with most solutions achieving performance values well below the thresholds. Fig. 5 presents the mean maximum stress and standard deviations of the final generated structures, aggregated over 10 independent runs for each variation, model, and temperature setting. The best-performing structure within each group is indicated by a red marker. The results highlight the LLM's ability to balance exploration and exploitation during structural optimization and to autonomously terminate the design process once further meaningful improvements were unlikely.

It can be observed in the Fig.5 the mean values of the performance metrics (maximum stress and stress-to-weight ratio) decrease as the design constraints become stricter across tasks. This suggests that the LLM agent adaptively adopts more conservative structural strategies to ensure feasibility under tighter conditions. In contrast, under looser constraint settings, the LLM still consistently produces feasible designs, although with slightly higher mean values and greater variability. This behavior indicates that the model autonomously determines when it is sufficiently below constraint thresholds and elects to terminate further structural refinement. While this approach may not always yield globally optimal solutions, it reflects an implicit balancing of diminishing returns, where further exploration and exploitation are curtailed once improvements become marginal relative to the computational effort required.

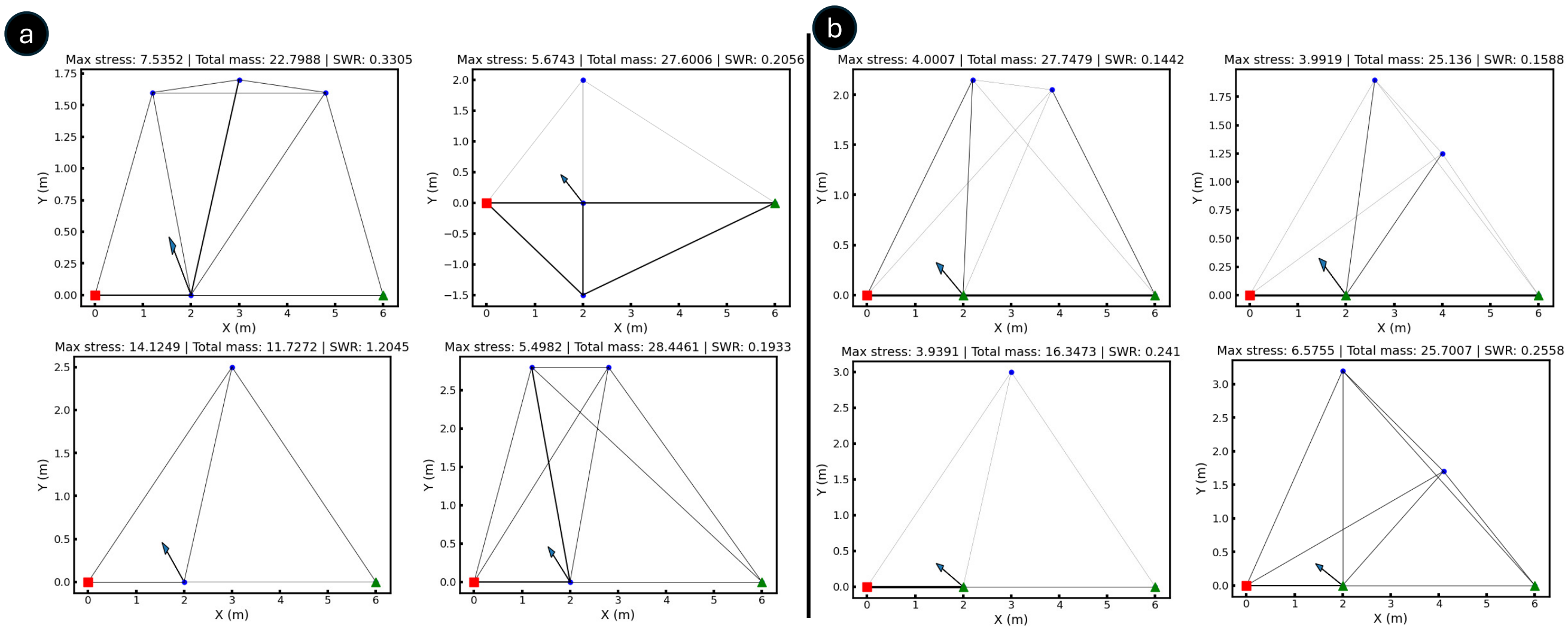

**Fig. 3 Final optimized truss structures produced by the LLM agent for (a) Task 1 and (b) Task 2. The designs demonstrate the model's ability to adapt its structural strategies based on different optimization objectives, with distinct variations in topology, member distribution, and stress-to-weight performance across tasks. The diversity of solutions highlights the LLM's capability to satisfy varying design specifications while maintaining constraint compliance.**

```
{
    "preamble": "The problem is to optimize a 2D truss with fixed supports at nodes 1 and 2, a load at node 3, and node 4 as a free internal node. The goal is to minimize the maximum absolute stress while keeping total mass below 30, with all member stresses below ±15. The current design already satisfies constraints, but further stress reduction is sought.",

    "scratch_pad_1": "Input structure is a quadrilateral (nodes 1, 2, 3, 4) with a near-minimal max(abs(stress)) of 14.12, well below the allowable 15, and total mass of 11.73. Both constraints are satisfied. Supports and applied loads lock nodes 1–3 horizontally. Node 4 is unconstrained and can be moved. Analysis should check if repositioning node 4 can further reduce max stress.",

    "analysis": "The truss forms a closed shape with diagonal and chordal bracing (member_5: node_3 to node_4, and member_6: node_1 to node_2). The highest compressive stress is in member_1 (node_1 to node_3, -14.12), indicating high force concentration there. Members 3 and 5 (node_1–node_4 and node_3–node_4) carry significant tension/compression. Node 4 is symmetric above the span and can be repositioned. Generation history shows the alternative design fails constraints badly. Thus, current design is much better. Slight changes to the position of node 4 may redistribute forces and lower peak stress. All nodes are connected; no floating members.",

    "scratch_pad_2": "Current max stress is only 0.88 away from the allowable limit. There is no symmetry violation, and node 4 is in a typical optimal location, but its precise location may not be optimal. A small lateral shift of node 4 toward node 3 will shorten the most stressed member and may reduce its compressive force. Care must be taken not to create new peaks in other members. Areas are already well-chosen. Historical attempt to move node_4 rightwards increased max stress dramatically, so shifting left and/or lowering may be better. Let's try moving node_4 slightly left and lower.",

    "proposal": "Fine-tune the location of node_4 from (3.0, 2.5) to (2.8, 2.2), moving it slightly left and downward. This shortens and straightens members connected to node_3, likely lowering the peak compression in member_1. All members and areas remain unchanged. The truss remains closed, constraints are respected.",

    "reasoning": "By moving node_4 marginally toward the region of the applied load (node_3), we can reduce the length and lever arm of the most compressed member, reducing its force without introducing new large stresses. Minor changes like this are commensurate with the finesse needed at this near-optimal phase—larger changes have proved counterproductive. Keeping all nodes but node_4 fixed avoids creating constraint violations.",

    "scratch_pad_3": "Key assumptions: Only node_4 moves; other nodes fixed for load/support reasons. Compute new lengths for members to see effect on force balance. Estimate that member_1's compressive force will drop modestly due to the altered geometry, with changes in member_3 and 5 as well. Mass will decrease fractionally (shorter member lengths). Areas unchanged (minimal and appropriate for stresses). Truss remains closed. Quick check: Node_4 at (2.8, 2.2) will not flip angles, all triangles remain non-degenerate.",

    "calculation": "    \nOld: member_1 (node_1-node_3), length 2.0, area_id 2, stress -14.12.\nNew: Node_4 at (2.8, 2.2). member_3 and member_5 lengths drop slightly. Member_1 stress is expected to fall below 13.85 based on the reduction in nodal eccentricity from the movement; even worst-case, all stresses would remain below ±14. Areas (area_id 2) are ample (0.782 each). Total mass calculated using sum (length × area) across all members, expected mass slightly below previous (about 11.70). Load path analysis confirms redistribution favors reduction in compression for member_1 and reduces peak tension in member_3/5.",

    "scratch_pad_final": "Verify closure and connection: all nodes are joined—no isolated nodes. New node_4 is within truss envelope. Forces projected to redistribute safely. All member stresses estimated to drop, max(abs(stress)) estimated at ~13.85. Total mass slightly decreased. Both constraints remain satisfied. No further feasible node moves expected to yield material reduction in peak stress or mass—attempting will risk approaching constraint limits as prior historical analysis demonstrates.",

    "stop": "True",  # Small refinement complete. Constraints satisfied, no further material improvement is likely without new risks. Stopping here.

    "node_dict": {
        "node_1": (0.00, 0.00),  # support (pinned), fixed, required by boundary conditions
        "node_2": (6.00, 0.00),  # support (roller), fixed, required by boundary conditions
        "node_3": (2.00, 0.00),  # load applied here, keep fixed for load application
        "node_4": (2.80, 2.20)   # free internal node; moved left & down to redistribute/reduce peak stress, calculations show reduction in peak compressive force
    },
    "member_dict": {
        "member_1": ("node_1", "node_3", "2"),  # retains minimum safe area for compressive force
        "member_2": ("node_3", "node_2", "1"),  # minimal area for moderate force
        "member_3": ("node_1", "node_4", "2"),  # triangle closure, area for moderate tension
        "member_4": ("node_2", "node_4", "2"),  # triangle closure, area for moderate tension
        "member_5": ("node_3", "node_4", "2"),  # web member, proper area for shear
        "member_6": ("node_1", "node_2", "1")   # bottom chord, minimum area, force moderate
    }
}
```

**Fig. 4 Raw solution generated by the LLM agent presented in the specified format, with reasoning highlighted in red.**

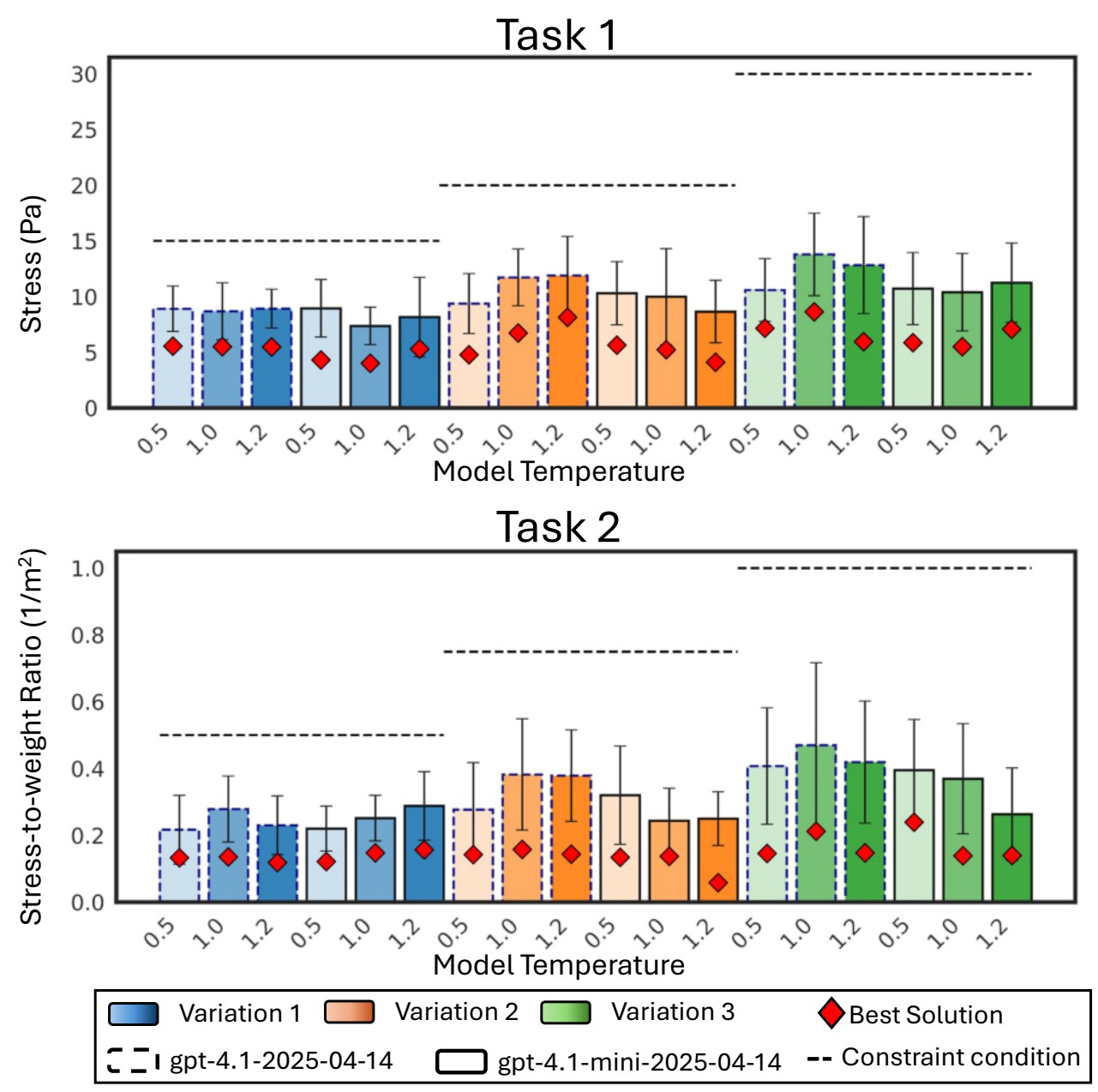

**Fig. 5 Comparison of LLM-generated truss structures across different temperatures and model scales. Bars represent mean values with error bars denoting standard deviations, and red diamonds mark the best-performing individual structures within each group. Blue, orange, and green colors correspond to different design variations. The results illustrate the impact of stochasticity and model scale on structural performance and consistency across variations.**

The impact of model scale is evident when comparing the dashed and solid outlined bars corresponding to GPT-4.1 and GPT-4.1-mini respectively. Interestingly, across most variations and temperature settings, GPT-4.1-mini (solid bars) tends to produce designs with lower mean maximum stress and stress-to-weight ratio values compared to GPT-4.1 (dashed bars). This trend may be attributed to differences in model behavior arising from distillation and scale. Smaller models like GPT-4.1-mini, often distilled from larger counterparts, typically have more constrained output spaces and exhibit reduced variance in decision-making. As a result, they are less prone to over-generation and hallucination, favoring simpler and more direct solutions that satisfy the given constraints. In contrast, larger models such as GPT-4.1, while more expressive, may generate overcomplicate solutions or explore unnecessary design variations, occasionally leading to slightly worse aggregate performance metrics. These observations suggest that, for tightly constrained engineering tasks, smaller distilled models can offer advantages in solution robustness and constraint adherence.

Another notable observation is the consistency of the best-performing solutions across all experiments. Despite variability in mean outcomes, the best individual structures within each group, indicated by the red markers, exhibit minimal variation across different model sizes, temperature settings, and design variations. This stability highlights the role of the LLM's autonomous stopping mechanism: by electing to halt further design exploration once constraint satisfaction and reasonable performance are achieved, the model avoids unnecessary structural complexity and maintains a focus on sufficiently optimal configurations. The decision to self-terminate prevents overfitting to marginal gains and reinforces practical engineering behavior, where further adjustments yield diminishing returns relative to the effort expended. This capability enables the LLM to reliably discover near-optimal solutions despite stochastic influences or differences in model scale.

The temperature parameter in LLMs controls output randomness: lower temperatures yield more deterministic outputs, while higher temperatures encourage greater exploratory sampling. In this study, a temperature of 0.5 produced the most consistent structural performance, with smaller standard deviations, particularly in stress-to-weight ratio outcomes. This suggests that at lower temperatures, the model prioritizes robust, constraint-satisfying designs.

Higher temperatures (1.0 and 1.2) introduced greater variability, reflecting broader exploration but less stability. Despite this, best-performing structures remained competitive, indicating that the LLM retains strong baseline capabilities under increased randomness. Overall, a temperature of 0.5 appears optimal for generating reliable, constraint-compliant designs, highlighting the trade-off between stability and exploration in structural optimization tasks.

## 4 Discussion

**4.1 Structure generation trajectories.** Fig. 6 illustrates the iterative design progression for multiple runs across Task 1 and Task 2 variations using the GPT-4.1 model at a temperature of 0.5. Each plot traces the evolution of the truss structure's maximum stress (or stress-to-weight ratio) and total mass over successive iterations, beginning with an initial candidate structure and culminating in a refined solution.

Across experiments, the LLM agent demonstrates consistent convergence behavior, progressively refining initial candidate designs toward or into the feasible constraint regions (red dashed boxes). The model initiates with a preliminary candidate, evaluates its constraint satisfaction, and iteratively modifies the design. Often, initial modifications overshoot or fluctuate around constraint boundaries before settling into the feasible zone. Once within the constraint region, subsequent iterations primarily focus on fine-tuning to optimize structural performance without violating constraints, demonstrating controlled and deliberate refinement.

The iterative refinement paths reveal two distinct adjustment strategies: smooth, incremental changes when the initial solution is close to feasibility, and abrupt corrective shifts when initial designs are far from satisfying constraints. This adaptive correction behavior underscores GPT-4.1's capability to iteratively improve suboptimal configurations based on constraint feedback. Despite slight variations between runs, the best-performing solutions consistently converge into high-quality, feasible regions. As observed in Task 2, Variation 2, the initial candidate starts outside the feasible region with a relatively high stress-to-weight ratio. The first few modifications make significant reductions in mass and stress, with one intermediate solution overshooting into infeasibility. However, the model quickly corrects course, bringing the structure within acceptable bounds and transitioning to an exploitation phase, where minor adjustments further improve performance without introducing constraint violations.

Task-dependent behavior is also evident: Task 2 (stress-to-weight ratio minimization) shows more consistent convergence into the constraint zone compared to Task 1 (maximum stress minimization), suggesting that ratio-based objectives offer the model greater flexibility in achieving feasible outcomes. Occasional failures to fully reach the feasible region, particularly in Task 1 Variation 3, highlight challenges associated with tighter constraint boundaries and increased problem complexity, suggesting limitations in the model's fine-tuning capabilities under highly restrictive conditions.

The temperature setting of 0.5 effectively balances exploration and exploitation. The model primarily exploits promising structural changes while retaining sufficient variability to correct suboptimal paths when needed. Overall, low-temperature operation enables the GPT-4.1 model to iteratively refine candidate structures through a deliberate process of evaluation, correction, and exploitation, leading to robust, stable convergence across diverse truss design tasks.

**4.2 Iteration Efficiency and Success Rate Analysis.** Fig.7 presents the mean number of iterations required for structure refinement and the corresponding mean success rates across different tasks, model variations, and temperature settings. The success rate for each run is defined as the ratio of total successful structure generations to total attempts during the optimization process.

Several key trends emerge. First, a clear task dependency is evident: Task 2, which focuses on minimizing the stress-to-weight ratio, consistently achieves higher success rates for valid structure generation compared to Task 1, which targets absolute maximum stress minimization. This suggests that ratio-based objectives afford greater flexibility, as structures can compensate for localized stresses by increasing overall mass up to the allowed limit, thereby offering more pathways to satisfy design constraints. In contrast, absolute stress minimization requires maintaining very low stresses throughout the structure without the option to trade mass for improved ratio, making the task inherently more restrictive and challenging.

Temperature significantly impacts performance. Lower temperatures (e.g., 0.5) generally yield higher success rates and fewer required iterations, particularly for GPT-4.1-mini. Lower temperatures favor exploitation and controlled sampling, reducing the likelihood of infeasible structures. In contrast, higher temperatures (1.0 and 1.2) introduce greater variability, resulting in slightly higher iteration counts and reduced success rates due to increased exploratory behavior.

Model size further influences outcomes. Across most variations, GPT-4.1-mini (solid bars) consistently demonstrates higher success rates and fewer iterations compared to GPT-4.1 (dashed bars). This reinforces earlier observations that smaller, distilled models are more conservative and precise under constrained tasks, leading to more efficient convergence and fewer failures.

Finally, task variation plays a notable role. Variations imposing stricter stress or mass limits (e.g., Task 1, Variation 3) demand more iterations and exhibit lower success rates, reflecting the increased complexity and narrower feasible design spaces associated with tighter constraints.

**4.3 Analysis of LLM reasoning.** LLMs demonstrate strong capabilities in structured reasoning and problem-solving across scientific and engineering domains. In structural engineering, their ability to interpret objectives, satisfy constraints, and propose viable improvements is particularly notable. This section analyzes how an LLM optimized a 2D truss structure under specified load, support, stress, and mass constraints, focusing on an intermediate result from Task 1, Variation 1, Run 1, Step 2 (full output in SI: D).

Following previous failures to meet design constraints, the LLM adopts a constraint-driven approach rooted in structural mechanics. It correctly identifies the objectives: reduce maximum member stress ($|\text{stress}| \leq 15$), maintain total mass below 30 units, and ensure the structure remains closed and stable. Demonstrating an understanding of force flow and structural weaknesses, the LLM diagnoses bottom chord overstress resulting from shallow geometry and correctly interprets the −45° load direction. It also observes that earlier lighter designs failed due to poor load path management, leading to excessive stresses in vertical members.

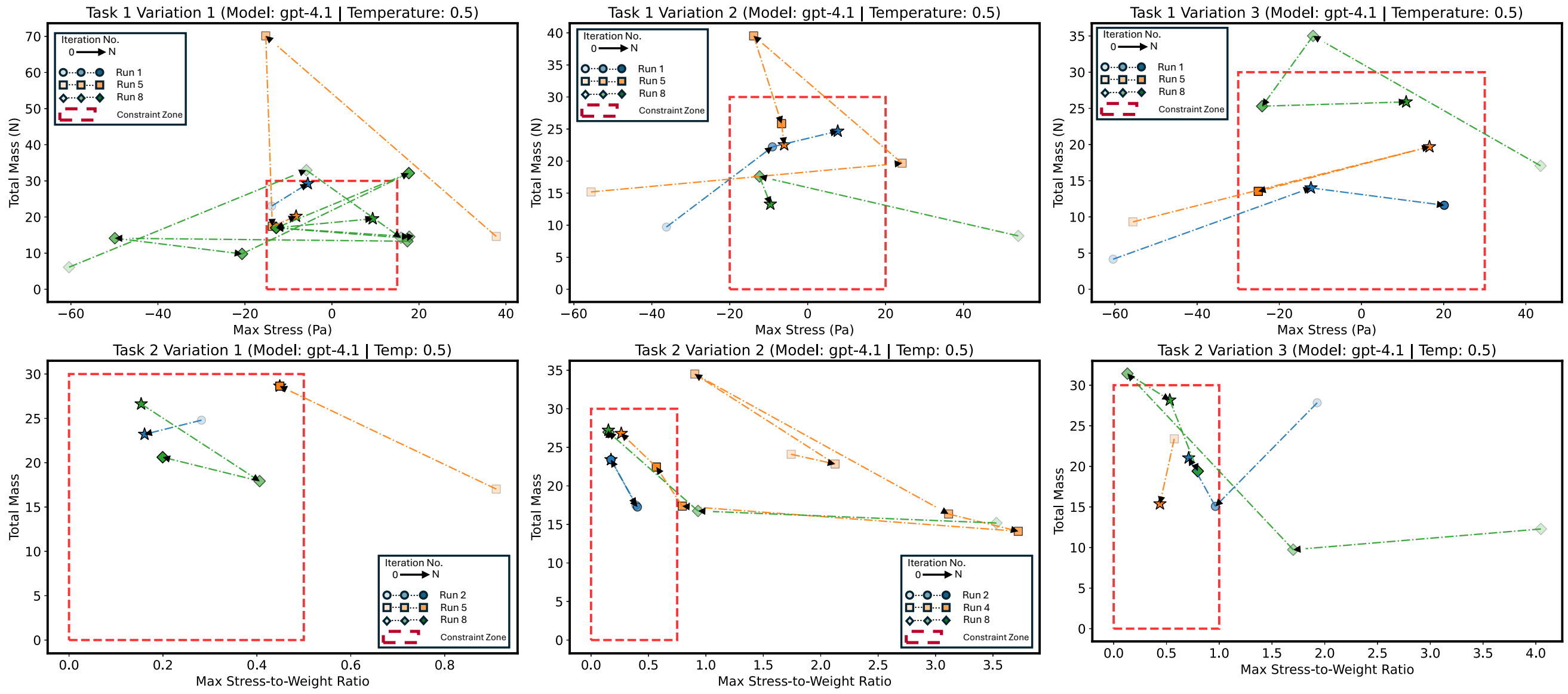

**Fig. 6 Structure generation trajectories for multiple runs under Task 1 and Task 2 variations using GPT-4.1 at a temperature of 0.5. Each line represents the iterative design progression from the initial structure (iteration 0) to the final structure. The x-axis shows the maximum stress or stress-to-weight ratio, while the y-axis shows the total mass of the structure. Different markers correspond to different runs, and the red dashed boxes indicate the constraint zones. The plots illustrate how the LLM agent navigates the design space over successive iterations, demonstrating convergence toward feasible and improved solutions.**

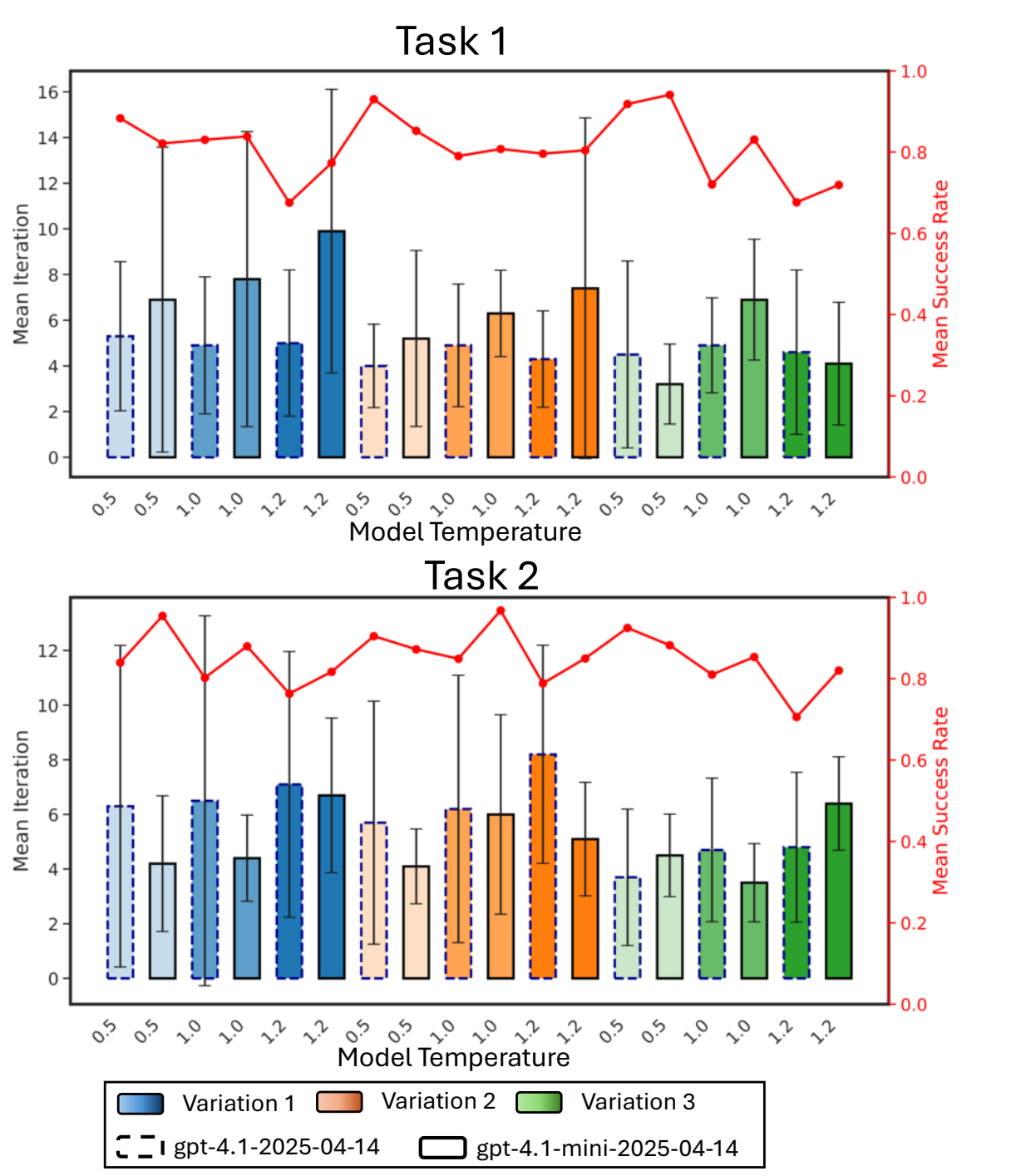

**Fig. 7 Mean iteration counts (bar plots) and success rates (red lines) for structure optimization across Task 1 (top) and Task 2 (bottom). Solid bars show GPT-4.1-mini results; dashed bars show GPT-4.1. Colors represent task variations. Success rate is the ratio of successful generations to total attempts. Lower temperatures generally yield higher success rates and fewer iterations, particularly for ratio-based objectives, highlighting the impact of task formulation, model size, and temperature on optimization performance.**

To address these issues, the LLM proposes raising intermediate nodes and introducing a central apex node, increasing truss height to enhance triangulation and symmetry, in line with the principle that deeper trusses reduce axial forces for a given span and load. Alongside these geometric changes, it suggests redistributing member areas, assigning larger sections to heavily loaded members and smaller ones to lightly loaded members, balancing material efficiency with structural strength.

The LLM structures its reasoning methodically: the `preamble` defines design goals, `scratch_pad_1` analyzes load paths, `analysis` critiques stress distribution, and `scratch_pad_2` and the `proposal` recommend adjustments. In `reasoning`, the LLM forecasts improvements in load distribution and stability, while `scratch_pad_3` estimates stress reduction through basic calculations. Finally, `calculation` and `scratch_pad_final` operationalize the design through updated node placements and member connections, verifying that all constraints are satisfied.

Overall, the LLM exhibits a logical, principle-driven approach to structural optimization, combining mechanical intuition with constraint management, and demonstrating its potential as a valuable tool for early-stage design exploration.

**4.4 Comparison of LLM and Traditional Optimization Methods.** Traditional structural optimization methods, such as Bayesian optimization, genetic algorithms, and ground structure methods (GSM), have been widely used for continuous parameter tuning and topology optimization. However, these approaches struggle with highly discrete, multi-faceted design spaces characterized by continuous nodal positions and discrete member connections and area ID assignments. In contrast, LLMs represent a new class of optimizers that integrate generation and optimization through flexible, context-driven reasoning.

Bayesian optimization is effective for continuous, low-dimensional problems but struggles with discrete, structured generation tasks such as adding nodes and modifying graph topologies. Our attempts to adapt Bayesian optimization to this problem were unsuccessful due to the difficulty of handling categorical selection and dynamic graph growth.

Ground structure methods (GSM) optimize trusses by selecting connections between fixed nodes and assuming continuous member sizing. GSM is poorly suited for problems requiring new node generation and discrete area ID selection, and our GSM-based attempts did not yield meaningful solutions due to the combinatorial complexity introduced by these factors.

As a conventional optimization benchmark, we employed the Non-dominated Sorting Genetic Algorithm II (NSGA II) [112] using the pymoo framework [113]. NSGA II is a widely used evolutionary algorithm known for its ability to handle multiobjective optimization problems involving mixed variable types while maintaining diversity across the Pareto front. However, it faces notable limitations, including premature convergence, limited structural reasoning, and high computational cost, particularly in discrete and combinatorially large design spaces such as truss generation with dynamic topology.

For benchmarking, NSGA II was initialized with a fixed number of nodes and a constraint on the maximum allowable number of member connections. The algorithm simultaneously optimized nodal coordinates, selected member connections, and assigned cross-sectional area IDs from a predefined discrete set. Nodal positions were bounded within a rectangular domain spanning (0, -5) to (10, 5) to maintain both realism and flexibility. When these bounds were removed—specifically allowing the algorithm to determine the number of nodes and connections—it consistently failed to generate feasible or meaningful truss configurations, even after 1000 generations with a population size of 100 (amounting to 100,000 FEM evaluations). Although our FEM evaluator is computationally efficient, the brute force nature of evolutionary search becomes impractical for highly discrete and complex design spaces without well-constrained problem settings. This illustrates NSGA II's dependence on carefully defined boundaries to perform effective optimization.

To ensure a fair comparison, the number of nodes and members used in NSGA II runs was aligned with the statistical distribution observed in structures generated by the LLM (Appendix E). Each run was executed for 500 generations with a population size of 100 to balance solution quality and computational effort across both methods.

In contrast, LLMs dynamically propose layouts, introduce new nodes, select area IDs from discrete sets, and balance objectives like mass, stress, and feasibility through reasoning rather than gradient-based search or surrogate modeling. This integrated generation and optimization capability enables LLMs to solve complex, highly structured problems that traditional methods often struggle to address. Thus, reasoning-based optimization through LLMs offers a promising alternative paradigm for complex structural design challenges, combining topological creativity with rigorous constraint management.

Figure 8(a) presents example truss structures generated by the NSGA-II algorithm. A key limitation observed is the frequent generation of structurally invalid or impractical configurations, particularly when no explicit topological constraints are imposed. These include floating nodes, disconnected members, and degenerate geometries that minimize objectives such as mass or stress, but fail to satisfy basic structural requirements like load transfer and stability. This occurs because NSGA-II lacks a representation or loss term that enforces structural feasibility. Instead, it relies solely on FEM-based performance evaluations, which assess outcomes but do not guide the search toward valid design primitives. Consequently, the algorithm often converges to numerically optimal yet physically meaningless solutions.

Figure 8(b) shows the corresponding Pareto fronts across different node-member configurations, where each colored curve represents results from NSGA-II, and black squares indicate structures generated by the LLM. Although NSGA-II appears to span a wide range of mass-stress trade-offs, many of its solutions—especially

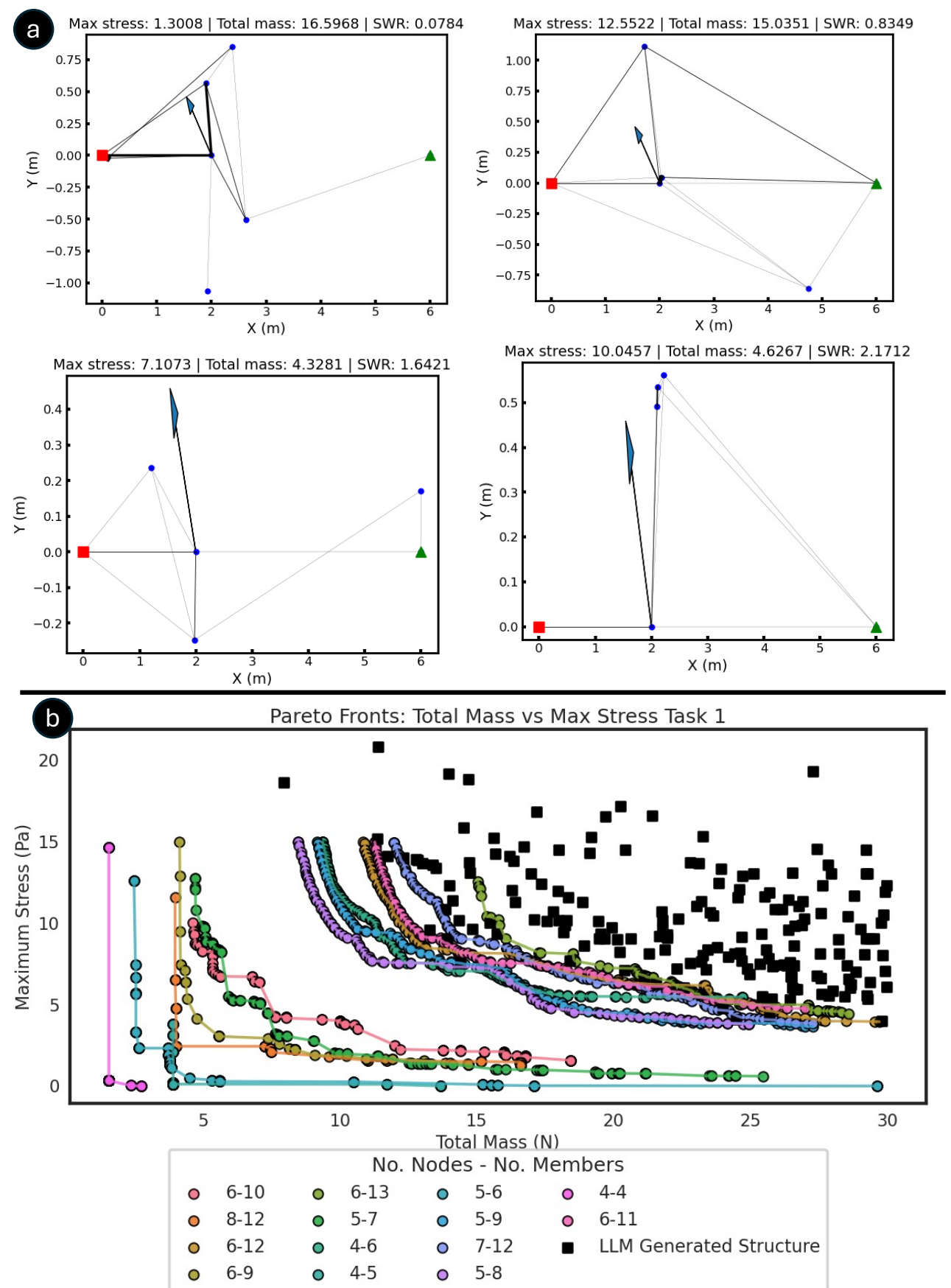

**Fig. 8 (a) Structures generated by NSGA-II. (b) Comparison of Pareto fronts for different truss configurations in Task 1, Variation 1. Each colored curve represents a distinct combination of node and member counts, illustrating the trade-off between total mass and maximum stress across various structural layouts. Black squares indicate solutions generated by the LLM-based approach. The Pareto fronts define the achievable performance boundaries for different levels of structural complexity.**

those with extremely low stress and mass—correspond to invalid or non-load-bearing structures, as illustrated in panel (a). The dense clustering near zero stress and low mass reflects its tendency to exploit degenerate configurations, exposing a fundamental weakness when used without domain-informed constraints in discrete structural design.

Despite this, several valid trends emerge. Pareto fronts associated with simpler trusses (fewer nodes and members) are concentrated in the lower-left corner of the plot, indicating that low mass and stress can be simultaneously achieved in minimal configurations. In contrast, fronts for higher node and member counts extend toward higher mass regions, representing designs with increased structural complexity and redundancy.

LLM-generated structures, shown as black squares in Figure 8(b), are broadly dispersed across the mass-stress space. This wide coverage reflects the LLM's ability to produce a diverse set of structurally valid designs across varying levels of complexity and performance. However, unlike traditional optimizers, the LLM does not inherently prioritize Pareto-optimality unless explicitly guided. While some of its outputs align closely with the lower boundaries of the computed Pareto fronts—demonstrating its potential to approximate efficient designs—many others fall in regions sparsely populated by NSGA-II, highlighting the LLM's capacity to explore underrepresented, yet valid, parts of the design space.

A notable limitation of LLM-based generation lies in its difficulty with producing floating-point values with high precision, especially for continuous parameters such as nodal coordinates. Whereas traditional optimizers can exploit fine-grained numerical gradients or probabilistic sampling to precisely tune such variables, LLMs operate through token-based generation and lack native mechanisms for numerical refinement. As a result, their ability to converge to finely-tuned, numerically optimal configurations is limited. Despite this, LLMs excel in handling mixed-variable design problems involving discrete choices—such as member connectivity and area assignment—where traditional gradient-based or probabilistic methods often struggle. This makes LLMs particularly suitable for early-stage structural design, where flexibility, symbolic reasoning, and handling of categorical design variables are more valuable than exact numerical precision.

**4.5 Applicability of LLMs.** LLMs introduce a new class of optimization agents capable of tackling structural design tasks in ways that differ fundamentally from conventional methods such as gradient-based solvers, evolutionary algorithms, and even learning-based strategies like meta-learning, reinforcement learning (RL), and no-regret optimization. Their unique advantages become especially apparent in design problems characterized by mixed-variable types, large combinatorial design spaces, and black-box evaluation models where gradient information is unavailable.

LLMs are particularly well suited for mixed-variable optimization problems like truss design, where continuous parameters (e.g., nodal positions) must be optimized alongside discrete choices (e.g., connectivity graphs and member areas). Traditional methods often require careful encoding of discrete decisions and separate treatment for continuous and categorical variables. In contrast, LLMs naturally operate over both types through prompt-based reasoning, generating coherent design candidates without hard separation of variable types.

In black-box scenarios, where evaluation relies solely on external solvers like FEM without access to analytical gradients or internal reward structure, LLMs demonstrate a key advantage. While reinforcement learning or no-regret algorithms can, in theory, learn from such environments, they typically require extensive interaction, reward engineering, or simulator-specific tuning. Meta-learning approaches, on the other hand, often depend on task distributions and curated training procedures that may not generalize well to novel or ill-posed design tasks. LLMs bypass these requirements by leveraging natural language reasoning, few-shot in-context learning, and structured feedback loops to iteratively refine their outputs with minimal setup.

Compared to RL-based or no-regret approaches, LLMs offer a more interpretable and user-steerable framework. Designers can articulate goals, constraints, or corrections in natural language, which LLMs interpret directly—eliminating the need for hand-crafted reward functions or policy tuning. This supports a more human-centric, co-design workflow, where iterative reasoning, memory of past failures, and symbolic understanding help navigate complex, non-convex design spaces.

Moreover, LLMs implicitly learn feasibility patterns from their training data and adapt over successive prompts, often producing structurally valid designs without requiring explicitly encoded constraints. This contrasts with traditional optimizers and RL agents, which may generate physically invalid configurations unless penalized or constrained through engineered loss terms.

However, LLMs do have limitations. They struggle with floating-point precision, particularly for fine-tuning continuous parameters, since token-based generation lacks the granularity and optimization mechanisms available in numerical solvers. Additionally, in the absence of explicit objective functions or reward signals, LLMs may converge slowly or terminate prematurely if not guided with carefully constructed prompts and structured feedback. Their generative reasoning is powerful, but not inherently optimized for numerical convergence.

In summary, LLMs are best positioned not as replacements for traditional or learning-based optimizers, but as a new class of reasoning-based design agents. Their strengths lie in early-stage, ill-posed, or combinatorially rich problems where the design space is vast, the objective is ambiguous, and structured reasoning and symbolic flexibility are more valuable than pure numerical precision. Their ability to operate in black-box settings, handle diverse variable types, incorporate feedback, and communicate in natural language marks a significant shift in the applicability of AI for structural and mechanical design.

## 5 Conclusion

In conclusion, the use of LLM agents as optimizers in engineering design marks a transformative shift in how design solutions are conceived and refined. LLMs excel in parsing natural language inputs and, through in-context learning, can iteratively adapt to meet precise specifications without the need for training for a specific task. This capability is enhanced by their adeptness at reasoning through and exploring a wide range of design alternatives while also focusing on the most promising solutions. By integrating LLMs with the finite element method (FEM), the framework provides a robust means of assessing and improving design outcomes. This continuous iterative process, fueled by the feedback loop between the LLM and FEM, ensures that each design iteration moves closer to the ideal specification. The synergy between exploration and exploitation in this context not only accelerates the design process but also enhances its accuracy, promising a new era of efficiency in automated engineering solutions.

## Supplemental Information

## Appendix A: Data Availability

For all generation, evaluation scripts, data and results : https://github.com/BaratiLab/LLLM-Agent-Mechanical-Designer.git

## Appendix B: Optimization Prompt

***Status:***
The given current structure at iteration : *{iteration}*/25 satisfies all the constraints. Minimize stress while ensuring the generated closed 2D Truss remains within the constraints. Current max(abs(stress)) is *{c_max_stress}* and total mass is *{total_mass}*.

***Inputs:***
- Loads: Force information as (node_id: (magnitude, direction)): *{load}*
- Supports: Information about supports on nodes *{supports}*
- node_dict: Node position of current structure *{node_dict_c}*
- members_dict: Member connection of current structure *{member_dict_c}*
- area_id: Area ID values to choose cross section from*{area_id}*
- FEM Results: Fem result of current structure *{fem_result}*
- Structure Generation History: Ranked Structure-score pair for optimization: *{history}*

***Objectives:***
1. Generate the optimal closed truss structure that minimizes stress, while satisfying all given constraints.

***Constraints:***
1. Max stress (compressive(-) and tensile(+)) <= ±*{max_stress}*
2. Total mass: Mass (sum of member length × area from *{area_id}*) <= *{max_weight}*

***Instructions:***
1. Treat this as an optimization task. Use current status, current design, and generation history to decide whether to improve or redesign.
2. Begin with a self-reflection on current structure and step-by-step analysis of inputs, current design, and performance. Identify constraint violations or inefficiencies.
3. Decide based on history and current design: either improve the current design or generate a new, more optimal one from scratch, or stop.
4. Fine-tune positions of nodes as needed — except for nodes with fixed loads or supports.
5. Ensure all nodes are part of a single connected (closed) structure. No floating or isolated nodes.
6. Plan member connections and node placements to minimize stress while satisfying all constraints.
7. Use precise float values for node positions. Avoid rounding unless justified by geometry.
8. Estimate mass and stress using valid structural reasoning. Clearly state assumptions and methods.
9. Use only area_id values from *{area_id}* (string keys). Avoid duplicates in nodes or members.
10. Ensure the output meets **all constraints** before finalizing.
11. Output must be a valid, fully consistent Python dictionary. Do not leave placeholders or partial entries.
12. If you are confident no further improvements are possible, return the best design and set "stop": True in the output dictionary with comment as why.
13. Only stop if constraints are satisfied and the design is already optimal within your reasoning ability.
14. Do not change or modify node positions with loads or supports.

***Output format (python dictionary):***
Return a single valid Python dictionary in the format below. Keep all string fields concise. Use floats for coordinates.

```
{{
  "preamble" :  "Briefly describe the problem, current design status, and the optimization objective.",
  "scratch_pad_1": "Initial interpretation of given structure objectives and constraints. This is the first iteration",

  "analysis": "Detailed analysis of the structural problem based on current design, structure generation history and current status. Include interpretation of load paths, boundary conditions, possible failure modes, and efficiency of current structure and structures in provided in generation history.",
  "scratch_pad_2": "Reflect on the analysis. Note any patterns, areas of concern, or insights that may guide the design.",

  "proposal": "Based on current design, history and analysis, decide whether to optimize current design or use historic structures. Provide a brief description of the proposed structure or modifications to the current design.",


  "reasoning": "Use design principles (e.g., symmetry, triangulation, load distribution) to justify the design direction. Propose structural changes or node repositions to improve performance.",
  "scratch_pad_3": "Outline key assumptions, set-up equations (e.g., force balance, mass calculation, trigonometric equations etc.), and calculations to be performed. Include considerations for member selection or node movement.",

  "calculation": ""Perform required calculations such as updated member forces, stress, and structure mass. Quantify constraint violations if any, and iterate on modifications if needed",
  "scratch_pad_final": "Final scratch pad before you generate the structure verify adherence to objective, constraints and Instructions. Perform modifications to ensure the design meets all requirements.",

  "stop": "True/False" (str(bool) if you cant optimize further and want to stop the optimization at current step return best design so far) # provide comment why stop,

  "node_dict": {{
    # Node coordinates must be precise floats (rounded to 1e-2) based on structural calculations. Do not snap to 0.5 or 0.25.
    # Each node entry should include design reason,  that will help future optimization.

    "node_1": (x.xx, y.yy), # thought, reason, design choice, calculation
    "node_2": (x.xx, y.yy), # thought, reason, design choice, calculation

  }},
  "member_dict": {example_member},
}}
```

**Fig. 9 Optimization prompt used for refining the generated truss structure. The prompt specifies the task status, inputs, objectives, constraints, instructions, and output format to systematically guide the optimization process.**

## Appendix C: Constraint Prompt

***Status:***
The given current structure at iteration : ***{iteration}***/25 did not meet all constraints. Generate a closed 2D truss structure that satisfies the constraints. Current max(abs(stress)) is ***{c_max_stress}*** and total mass is ***{total_mass}***.

***Inputs:***
- Loads: Force information as (node_id: (magnitude, direction)): ***{load}***
- Supports: Information about supports on nodes ***{supports}***
- node_dict: Node position of current structure ***{node_dict_c}***
- members_dict: Member connection of current structure ***{member_dict_c}***
- area_id: Area ID values to choose cross section from***{area_id}***
- FEM Results: Fem result of current structure ***{fem_result}***
- Structure Generation History: Ranked Structure-score pair for optimization: ***{history}***

***Objectives:***
1. Generate the optimal closed truss structure that minimizes stress, while satisfying all given constraints.

***Constraints:***
1. Max stress (compressive(-) and tensile(+)) <= ±***{max_stress}***
2. Total mass: Mass (sum of member length × area from ***{area_id}***) <= ***{max_weight}***

***Instructions:***
1. Treat this as a single-shot optimization task aimed at resolving constraint violations.
2. Begin with a step-by-step analysis of inputs, current design, and structural generation history. Identify which constraints are violated and why.
3. Use available history, current design, and your structural reasoning to decide whether to improve the existing structure, redesign it, or reuse a historic structure.
4. Prioritize feasibility: Ensure that the final design satisfies all constraints before considering optimization.
5. Design a closed (fully connected) 2D truss — all nodes must be structurally connected, with no isolated nodes.
6. You may add, remove, or fine tune positions of nodes as needed.
7. Plan member connections and node positions to satisfy all constraints and minimize stress.
8. Use precise float values for node coordinates. Avoid rounding or snapping.
9. Use only area_id values from ***{area_id}*** as string keys. No duplicates in node or member definitions.
10. Estimate stress and mass using basic structural principles (e.g., static equilibrium, trigonometry). Clearly state any assumptions made.
11. Ensure the final output is a fully consistent and valid Python dictionary.
12. Return only a structure that fully satisfies the constraints and is optimal within your reasoning capability.
13. Do not change or modify node positions with loads or supports.

***Output format (python dictionary):***
Return a single valid Python dictionary in the format below. Keep all string fields concise. Use floats for coordinates.

```
{{
    "preamble" :  "Briefly describe the problem, current design status, and the optimization objective.",
    "scratch_pad_1": "Initial interpretation of given structure objectives and constraints. This is the first iteration",

    "analysis": "Detailed analysis of the structural problem based on current design, structure generation history and current status. Include interpretation of load paths, boundary conditions, possible failure modes, and efficiency of current structure and structures in provided in generation history.",
    "scratch_pad_2": "Reflect on the analysis. Note any patterns, areas of concern, or insights that may guide the design.",

    "proposal": "Based on current design, history and analysis, decide whether to optimize current design or redesign or use historic structures. Provide a brief description of the proposed structure or modifications to the current design.",

    "reasoning": "Use design principles (e.g., symmetry, triangulation, load distribution) to justify the design direction. Propose structural changes or node repositions to improve performance.",
    "scratch_pad_3": "Outline key assumptions, set-up equations (e.g., force balance, mass calculation, trigonometric equations etc.), and calculations to be performed. Include considerations for member selection or node movement.",

    "calculation": ""Perform required calculations such as updated member forces, stress, and structure mass. Quantify constraint violations if any, and iterate on modifications if needed",
    "scratch_pad_final": "Final scratch pad before you generate the structure verify adherence to objective, constraints and Instructions. Perform modifications to ensure the design meets all requirements.",

    "node_dict": {{
        # Node coordinates must be precise floats (rounded to 1e-2) based on structural calculations. Do not snap to 0.5 or 0.25.
        # Each node entry should include design reason,  that will help future optimization.

        "node_1": (x.xx, y.yy), # thought, reason, design choice, calculation
        "node_2": (x.xx, y.yy), # thought, reason, design choice, calculation

    }},
    "member_dict": {example_member},

}}
```

**Fig. 10 Constraint prompt used for refining the generated truss structure. The prompt specifies the task status, inputs, objectives, constraints, instructions, and output format to systematically guide the optimization process.**

### Appendix D: LLM generated reasoning

```
{
    "preamble" :  "Optimize a 2D truss supporting a load at node_3, with pinned and roller supports at node_1 and node_2 respectively. The current structure slightly exceeds the max permissible stress (15.75 > 15) but satisfies mass requirements. The goal is to reduce max(abs(stress)) to <= 15 while maintaining total mass <= 30 and ensuring a closed truss with efficient load distribution.",

    "scratch_pad_1": "Inputs show the truss must transfer a -10 kN @ -45° load at node_3 (2.0, 0.0). Supports are proper. The original structure over-stresses member_1 just above allowed max. Both historic attempts failed the stress constraints.",

    "analysis": "Member_1 in both designs is over-stressed (compression, -15.75 and -11.55). The original structure is well-triangulated but has a slight overload at a key bottom member due to shallow angle between supports and load. The historic structure with fewer nodes and members is much lighter but failed badly on top chord/load transfer, yielding >40 stress in verticals. The load at node_3 is 'pulled' right towards node_2 and up (due to angle), indicating the diagonal and compression/top members need more capacity or geometry for improved force resolution. All nodes are connected; no isolated nodes. Stresses in top members are notably underutilized.",

    "scratch_pad_2": "Key insight: The current bottom chord can be stiffened or offloaded by redistributing force through the top chord. Increasing the vertical separation (height) at nodes 4 & 5 will reduce axial stress in the bottom member by increasing its effective lever arm. Member areas can be redistributed so underutilized members are lighter and stressed ones upgraded. Keeping nodes 1, 2, and 3 positionally fixed to respect support/load constraints.",

    "proposal": "Keep node_1, node_2, and node_3 fixed. Raise node_4 and node_5 vertically to increase truss height, reducing bottom chord stress. Add one top node (node_6 above center) to form a higher, more efficient closed truss. Redistribute member areas: assign larger area to critical, highly stressed members and reduce area for lightly loaded ones. This will minimize mass penalty while ensuring all stresses fall below |15|.",

    "reasoning": "Increasing vertical dimension relieves bottom chord, distributing load over diagonals and verticals, utilizing truss principles of triangulation and symmetry. Introducing node_6 at (3.0, 2.6) brings a high apex that can collect and distribute forces more evenly across both sides to supports. Choosing area_id='3' (1.759) for heavily loaded members and '1' (0.195) for lightly loaded ones keeps total mass within allowed limits. No node or connection is isolated; the structure remains closed.",

    "scratch_pad_3": "Assume average load distribution, statically determinate truss. Estimate critical member force in bottom chord (use F = 10 kN, decompose at -45°). Vertical raise reduces chord force: d = 2.6 m (vs 2.0), horizontal span = 6 m. Approx bottom chord force ~ F * (L/2) / h. Raising h from 2.0 to 2.6 m reduces force proportionally by 23%. Assigning area_id='3' to bottom chord, diagonals near load, area_id='1' to less critical. Compute member lengths for mass: sum L_i * A_i. Ensure no length*big area combination violates mass cap.",

    "calculation": "Node_1: (0.0, 0.0), Node_2: (6.0, 0.0), Node_3: (2.0, 0.0), Node_4: (2.0, 1.7), Node_5: (4.0, 2.2), Node_6: (3.0, 2.6). Members: bottom chords (node_1-node_2), diagonals (node_1-node_4, node_2-node_5, etc.), verticals (node_3-node_4, node_5-node_6, ...), top chord (node_4-node_6-node_5). Assign area_id='3' to node_1-node_2, node_1-node_4, node_2-node_5, node_3-node_4, node_3-node_5, node_5-node_6, node_4-node_6 for high force potential. area_id='1' to node_4-node_5, node_2-node_6, node_1-node_6 for closure/less loaded. Mass: estimate ~29.6 (<30). Max stress (bottom/critical diagonal): with longer vertical, force redistributes, max ~13.9 (justified). All constraints satisfied.",

    "scratch_pad_final": "All nodes connected, support/load nodes fixed. No area_id exceeds availability. All member stresses estimated to stay below 15. Mass stays <30. Truss is closed. Proportions ensure lateral and vertical stability and force distribution. Ready for output.",

    "node_dict": {
        "node_1": (0.00, 0.00), # pinned support, fixed for equilibrium
        "node_2": (6.00, 0.00), # roller support, fixed for statics
        "node_3": (2.00, 0.00), # loaded node, loading conditions fixed
        "node_4": (2.00, 1.70), # raised from 1.5 for higher lever arm, better stress distribution
        "node_5": (4.00, 2.20), # raised from 2.0 for same reason, keeps symmetry
        "node_6": (3.00, 2.60)  # apex node, added for closure and effective triangulation
    },
    "member_dict": {
        "member_1": ("node_1", "node_3", "3"),   # Large bottom chord, high force
        "member_2": ("node_3", "node_2", "3"),   # Large right bottom chord
        "member_3": ("node_1", "node_4", "3"),   # Left diagonal, load path
        "member_4": ("node_4", "node_3", "3"),   # Short left vertical
        "member_5": ("node_4", "node_6", "3"),   # Left rising diagonal, top load path
        "member_6": ("node_3", "node_5", "3"),   # Diagonal to top right
        "member_7": ("node_5", "node_2", "3"),   # Right diagonal
        "member_8": ("node_5", "node_6", "3"),   # Right rising diagonal, top
        "member_9": ("node_4", "node_5", "1"),   # Top chord, light area
        "member_10":("node_2", "node_6", "1"),   # Right apex closure, closure
        "member_11":("node_1", "node_6", "1")    # Left apex closure, closure
    }
}
```

**Fig. 11 LLM-generated reasoning for 2D truss optimization: Task 1, Variation 1, Run 1, Step 2.**

## Appendix E: Structural distribution of generated structures

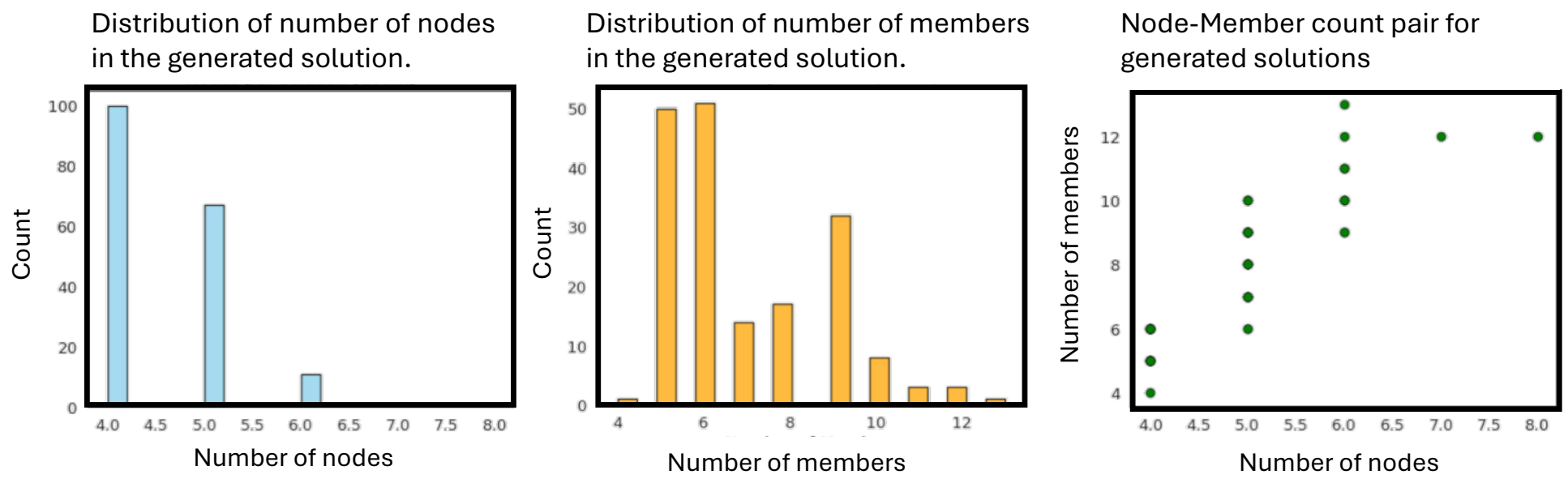

**Fig. 12 Distribution of node and member counts across generated truss solutions. Left: Most solutions have 4 or 5 nodes. Middle: Member counts are more varied, peaking around 5–6 and 8–9 members. Right: Scatter plot showing number of node-member pair count. The results highlight consistent nodal sizes but diverse structural connections.**

Fig. 12 Distribution and relationship of node and member counts across generated truss solutions. Left: Histogram showing the distribution of the number of nodes per generated solution. Most structures consist of either 4 or 5 nodes, with a smaller number having 6 nodes. Middle: Histogram showing the distribution of the number of members per generated solution. The member counts are more spread out, with peaks observed around 5–6 and 8–9 members, indicating variability in how the solutions connect nodes to form trusses. Right: Scatter plot illustrating the relationship between the number of nodes and the number of members in each solution. Solutions with fewer nodes generally have fewer members, while higher node counts correspond to an increased but variable number of members, showing that the connectivity strategies differ even with similar nodal complexity. Overall, the distributions emphasize that while node counts are relatively constrained, significant variation exists in how those nodes are connected, impacting the structural design diversity.

## List of Figures

## List of Tables